# Big data-driven prediction of airspace congestion

Samet Ayhan, Ítalo Romani de Oliveira, Glaucia Balvedi, Pablo Costas

Boeing Research & Technology

Alexandre Leite, Felipe C. F. de Azevedo

MWF Services

*Abstract* — Air Navigation Service Providers (ANSP) worldwide have been making a considerable effort for the development of a better method to measure and predict aircraft counts within a particular airspace, also referred to as airspace density. An accurate measurement and prediction of airspace density is crucial for a better managed airspace, both strategically and tactically, yielding a higher level of automation and thereby reducing the air traffic controller's workload. Although the prior approaches have been able to address the problem to some extent, data management and query processing of ever-increasing vast volume of air traffic data at high rates, for various analytics purposes such as predicting aircraft counts, still remains a challenge especially when only linear prediction models are used.

In this paper, we present a novel data management and prediction system that accurately predicts aircraft counts for a particular airspace sector within the National Airspace System (NAS). The incoming Traffic Flow Management (TFM) data is streaming, big, uncorrelated and noisy. In the preprocessing step, the system continuously processes the incoming raw data, reduces it to a compact size, and stores it in a NoSQL database, where it makes the data available for efficient query processing. In the prediction step, the system learns from historical trajectories and uses their segments to collect key features such as sector boundary crossings, weather parameters, and other air traffic data. The features are fed into various regression models, including linear, non-linear and ensemble models, and the best performing model is used for prediction. Evaluation on an extensive set of real track, weather, and air traffic data including boundary crossings in the U.S. verify that our system efficiently and accurately predicts aircraft counts in each airspace sector.

## I. Introduction to the Problem to be Solved

The last couple of years have been of recovery in the global air traffic, and we are approaching pre-pandemic levels of traffic. The FAA Office of Aviation Policy and Plans (APO) analysis of previous years showed that the cost of delayed flights peaked at $33.0 billion at 2019. Most of this cost was due to inefficiency in the National Airspace System (NAS), caused by many factors, among which the inability to accurately predict future states of airspace and airports, resulting in added costs to the airlines and passengers [1]. Notwithstanding, forecasts of the aviation industry estimate that the demand for passenger-km will roughly double between now and 2040 [2, 3], a challenge that becomes even bigger with the industry commitment to become net zero in carbon emissions by 2050 [4]. Thus, besides the use of alternative sources of energy, efficiency improvements are direly needed.

To tackle these challenges, the Air Traffic Control (ATC) services are in continuous improvement, within the broader implementation of advanced Air Traffic Management (ATM) concepts [5]. These modernization items include Trajectory-Based Operations (TBO), airspace redesign, dynamic resectorization, free-flight sectors, and other initiatives, all having the goal of improving the efficiency of the airspace. The expansion and deployment of these new ATM concepts and services should enable higher degrees of automation and predictability, yielding reduced levels of air traffic controller's workload. This paper presents a data-driven solution to predict airspace sector congestion and enable a more proactive management of the airspace capacity imbalances, thereby mitigating congestion and maintaining efficiency of the airspace system at a higher standard.

## II. Measuring Airspace Capacity and Congestion

Initiatives for measuring and improving airspace capacity are guided by the more general concepts of controller workload and airspace complexity [6]. Plenty of methods have been proposed to address the measurement and prediction of complexity for a particular airspace. Industry and partners led by the FAA and NASA proposed a metric that includes both traffic density (a count of aircraft in a volume of airspace) and traffic complexity (a count of potential aircraft conflicts in a volume of airspace), referred to as Dynamic Density (DD) [7]. Within a large effort initiated by the Radio Technical Commission for Aeronautics' (RTCA) Taskforce 3, researchers developed and validated several DD metrics to accurately predict sector density and complexity [8, 9, 10].

In order to use DD as a planning tool, it is necessary to project its behavior over the planning horizon. Hence, Sridhar et al. made an effort to predict dynamic density up to 20 minutes in advance [11]. Masalonis et al. evaluated potential metrics from past studies to see how well they predict DD at time horizons required for TFM decision support (up to 120 minutes) [12]. With a concretely defined measure of complexity, such as DD, the capacity of an airspace sector can be defined in terms of the maximum complexity acceptable. Timely and accurate prediction of imbalances between capacity and demand can help traffic managers to make assertive decisions and make the best use of airspace resources. However, the calculation of DD requires the collection of many variables [9] and, consequently, large expenses of resources, but there are simpler measures that can approximate it [13, 14].

Although these efforts in the aviation domain yielded a valuable DD metrics, a complete system aircraft trajectory big data management and analytics has not yet been proposed, so as to address the prediction of airspace sector densities, a crucial factor in computing DD. Besides, the time horizon for the prediction of airspace sector densities have been limited with 120 minutes. One of our previous works [15] was aimed at addressing aircraft big trajectory data management and analytics. However, it used an obsolete data model, called Aircraft Situation Display to Industry (ASDI), which the FAA has stopped supporting. It also used a Relational Database Management System (RDBMS) with a traditional database engine that suffered from inefficiency.

In the present work, we use up-to-date SWIM data sources and describe a novel method to process and maintain such type of big data, and with it, to compute and predict traffic congestion.

## II. BACKGROUND RESEARCH ON BIG DATA MANAGEMENT

The System Wide Information Management (SWIM) is the main data services architecture [16] supporting modern ATM, with many applications in the operational arena, as well for Research & Development [15, 17, 18, 19]. The current SWIM data services from the FAA [20] are developed to provide flight information services to a wide variety of aviation consumers, in standardized data formats. These services publish flight plans, aircraft positions, trajectories, and flow control messages, among others, from each of the 20 Air Route Traffic Control Centers (ARTCC) systems, and from the Air Traffic Control System Command Center (ATSCC). Although many messages contain a unique flight identifier, they are mostly uncorrelated, noisy, streaming at higher rates, and large in volume. For a period of 172 days between February and August 2018, the average size of single day collection of raw messages, from just one of the SWIM data feeds, the so-called TFMData [21], is 14.1 Gigabytes, or 1 Terabyte each 71 days . Clearly, for any analytics to be useful over this dataset, a data management solution is required.

Big trajectory data management and analytics remains an active area of research both in the data management and data mining domain as well as in the transportation domain. Several systems have been proposed in the data management and data mining domain thus far that could be used in this application area. BerlinMOD [22], PIST [23], and TrajStore [24] offer a similar storage and indexing technique using traditional database engines. Simba [25] and SpatialHadoop [26] enable distributed spatial analytics based on MapReduce programming model. Although they are capable of processing big trajectory data, they suffer from inefficiency due to high overhead rates of the underlying programming model. A column-oriented storage system such as SharkDB [27] offers an efficient query processing and analytics capabilities. However, due to complete in-memory approach it falls short when handling massive volumes of trajectory data. Unlike other systems, CloST [28], Elite [29], PARADASE [30], and the cloud based system of [31] offer a specific partitioning technique in distributed environments for a potential of higher query processing efficiency. UlTraMan [32] is another system aiming to deliver efficient query processing in a distributed environment, with more flexible partitioning approach. Nonetheless, it does not integrate the meta-table construction into global indexes. There are a few other systems that support distributed storage and computing by having integration with Apache Spark [33]. These systems include Apache Geode [34], Apache Ignite [35], IndexRDD [36], and SnappyData [37]. Although these systems leverage Apache Spark in support of efficient streaming, transactions, and interactive analytics, they do not provide flexible operations and optimizations for trajectory data analytics.

In fact none of the above systems is specifically developed for management and analytics of aircraft trajectory and flow big data. They are systems developed mostly with the purpose of handling trajectories and georeferenced data generated by mobile devices, with applications in ground transportation and consumer analytics. Adapting these systems to our specific use case in aviation domain would require a considerable amount of extra effort, if feasible at all. Hence, in response to these gaps, Boeing Research & Technology's Airspace Operational Efficiency (AOE) group has implemented a novel data management and analytics system using SWIM big data, presented in the next sections. This system was first published in [38], and here we explain it using an easier jargon and with an application case study.

## III. DATA META-PREPARATION SCHEME

Our client applications interact with databases that contain SWIM data collections, by means of microservices. For simplicity sake, we refer to these databases as SWIM databases, and there are two of them, a raw one and a pre-processed one. The raw SWIM database, referred to as R-SWIM, is fed with large numbers of messages published in SWIM, and is kept in daily collections, with the minimum length of days necessary for a flight lifecycle (5 days suffice). Each of these collections has an average size of 14.1 Gigabytes and no structuring, which make them very difficult to query. The Pre-processed SWIM database (P-SWIM) contains data extracted from R-SWIM but filtered, structured for queries and indexed. Both of them are managed by MongoDB servers [39].

The generation of P-SWIM collections from R-SWIM is performed by means of ad-hoc aggregations and Map-Reduce [40], which is a data processing paradigm for condensing large volumes of data into useful aggregated results, typical of MongoDB, which, in turn, has a built-in function to perform it. In the ad-hoc aggregation stages, intermediate sub-collections are generated in such a way that the amounts of data become progressively smaller. During this processing, data is grouped according to the known queries and there may be data type conversions (e.g. from string to numeric). Non-used fields are discarded, and only the useful ones are passed on to intermediate collections. These operations may be performed in succession, according to the data relationships implied by the queries. Once certain data properties are obtained, the intermediate collections can be further processed by the built-in map-reduce operation. This results in pre-processed collections that optimize the query performance to acceptable running times.

This preparation scheme was initially implemented in Python scripts that run continually, checking if there are updates in the R-SWIM collections, in order to propagate those updates to the P-SWIM collections. However, our rapid prototyping development demands frequent changes in the application use cases that may expand or alter the contents of P-SWIM. Consequently, pre-processing schemes for specific queries and applications are not enough, and we established a meta-preparation process in order to allow more generality, and this is depicted in Fig. 1.

As it can be seen in the figure, there are five steps in the meta-preparation process to produce P-SWIM collections with consumable features, according to the following description:

**Step 1**: The R-SWIM collections are not originally indexed. They could however be indexed by fields which are not interesting for the specific preparation batch. For this reason, in this step we check which indices are used and create only those that are useful for the specific preparation.

**Step 2**: R-SWIM is scanned in order to find all unique flight reference numbers, and the initial P-SWIM collections are created, containing the flight reference numbers. The idea is to concentrate information of a single flight in a single document in order to achieve an information dense dataset.

**Step 3**: The P-SWIM collections created in the previous step are now populated with the most recent data on specific messages, to avoid redundancy. A first level parsing is also performed in order to ensure well-structured data and avoid data type checks in subsequent steps. If any data type conversion is necessary (e.g. string to date time format), it is done at this step. The step is repeated for each message type applicable to the purpose of the current preparation batch.

**Step 4**: The data stored in P-SWIM collections is now able to be internally correlated. No access to other collections outside P-SWIM scope is expected at this stage. The correlation output is stored in the same P-SWIM collections created with single-field documents and updated with recent and parsed messages.

**Step 5**: A further correlation and feature extraction is performed on those collections generated in the previous step. The results now are dumped into a new set of P-SWIM collections that contain only the extracted features.

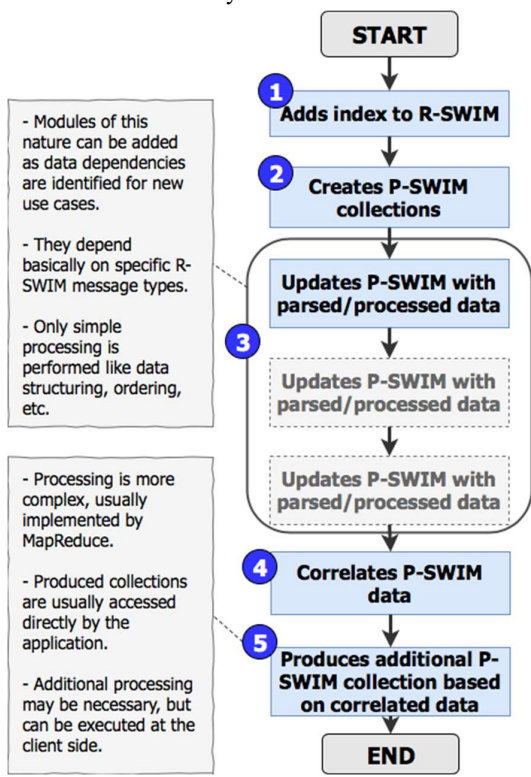

Fig. 1: Data meta-preparation scheme.

### IV. CASE STUDY ON PREDICTION OF AIRCRAFT COUNT

Although SWIM makes available a huge amount of data from at least 13 different producers [20], only TFMS (Traffic Flow Management System), a.k.a. TFMData [21], is in the scope of this prototype implementation, because it contains the flight data that can be correlated with airspace sector occupation. And, because of the difficulties involved in computing the DD, mentioned in Section II, we used aircraft count as the main measure of sector congestion, a measure that after all is highly correlated with DD [41].

#### A. Data preparation steps for aircraft count

Here we describe the application of the above meta-preparation scheme to produce collections containing aircraft count per airspace sector. The five steps in this specific case consist of:
1. Create the indices for flight reference and message type fields in R-SWIM collections.
2. Creates P-SWIM collection for a specific day with flight reference as a single field for each document.
3. Update the P-SWIM collection for that specific day with recent data parsed from R-SWIM collections with other message types. The search is performed in a range of five previous days where messages on that specific flight reference may be distributed in R-SWIM.
4. Correlate internally in a collection called DMS-A-DATE> to produce sector entry and exit times based on known departure and arrival times.
5. Perform sector count to produce the DMS-B-<DATE> P-SWIM collection whose documents are indexed by flight sector name. Thus, the amount of documents in this dumped collection has the same number of existing occurrences of unique flight sectors in all flights at DMS-A-<DATE>, which is around 1500 sectors.

The corresponding DMS-B collection is composed by documents uniquely named with airspace sector names, which occurred in the DMS-A documents on flights with valid departure times. Finally, a Map-Reduce routine can be executed on a smaller and filtered dataset to provide sector count with granularity of one minute. This is depicted in Fig. 2.

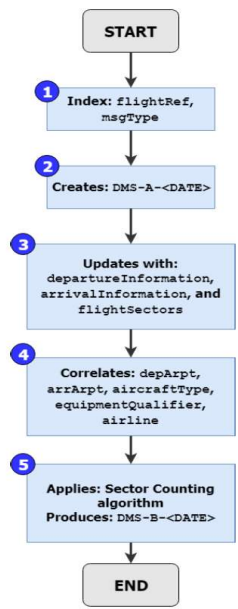

Fig. 2: Data preparation example for aircraft count per sector.

This process has the following advantages:
- Once a DMS-A collection is produced, it can be reused by different applications;
- Insertion of data from documents of other message types can be done at any time.

#### B. Details about the P-SWIM collections and the aircraft count case

This P-SWIM implementation is composed by the following types of collections: Status (ST), Flight Information (FI), Set Allocation (SA), Processed Information (PI), and Machine Learning (ML). These collections are populated daily by means of the meta-process illustrated in Fig. 1. FI has filtered data like most recent messages and parsed messages for individual flights. ST has information about the data storage in FI, since a lot of data flows into FI and partitioning them is very useful in terms of distributed disk space resources. SA expresses in an indexed way the location of documents with filtered flight information in the partitions of FI. PI stores processed data, which is the utmost objective of P-SWIM database. ML has an even higher abstraction level than PI, since its documents are machine learning models including estimators and other information necessary to speedup predictions.

This structure was prepared to afford scalability for processing and queries on big data. In our current framework, P-SWIM collections are populated as follows:

1. It is checked in ST which FI partition shall receive new entries of filtered data for individual flights.
2. Raw data in R-SWIM is filtered for single flights with multiple messages. A single filtered and parsed document is generated in the corresponding FI partition with this information.
3. SA contains documents which describe the assignment of each individual filtered flight to its containing partition.
4. PI receives processed information partitioned in daily collections out of filtered information spread through one or more FI partitions.
5. ML receives estimators/predictors trained with data available in PI.

In the current version of our systems, the fields with filtered information in a document of an FI collection are shown in Fig. 3.

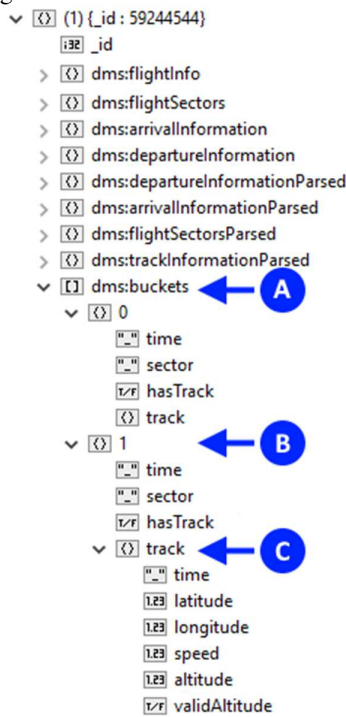

Fig. 3: A single document in the FI collection.

Those fields with the *Parsed* suffix are a parsed version of those fields with the same prefix. In this way, information that is normally distributed in R-SWIM gets consolidated in a dense way, since those messages with redundant information are not loaded into FI documents. The reason for keeping the non-parsed fields is that, once new data arrives, P-SWIM documents may have to be re-built and, in order to avoid re-processing old information from R-SWIM, the non-parsed fields already have that information at an advanced processing state.

The callouts of Fig. 3 indicate:
A. temporal series data after parsing and filtering;
B. an element (1-min time sample) of the temporal series array;
C. track information for other applications, inside the temporal series.

For sake of brevity, we focus the explanation on the role of only some fields and collections of Fig. 3 in the context of sector count case. The field `dms:buckets` stores a vector with entries like time (UTC), sector (sector name), hasTrack (flag for tracking content existence), and track (tracking data at that time sample shown in Fig. 3), grouping flight track data per period of time, which we call a bucket. `dms:buckets` is used to produce documents in PI.

Note that the generation of each bucket is a rather complex automated exploration with a series of MapReduce-like routines, since data must be adjusted to deal with issues like lack of expected records or missing fields, while keeping the consistency of its content. The trajectory of a flight with sector names as milestones is stored in `flightSectors` messages; however, these milestones are attached to entry times relative to the departure time. The departure times are present either in `departureInformation` message or `arrivalInformation` message. The latter message, as given, accumulates some of the flight history and one of its fields contains departure time and an additional qualifier field marking it as ACTUAL or ESTIMATED. Only departure times marked as ACTUAL are used in this case. Sometimes one of them does not exist or does not contain valid data; this requires an extra effort to extract the highest amount of information possible on each scenario. Sector name becomes an indexed field in PI documents with a list of flights in that sector ordered by 1-minute time buckets shown in Fig. 3.

*C. Regression model*

In order to use the data above explained for predicting aircraft count per sector, we need one or more regression models. Here we explain the regression model used in this case study.

This regression model is based on the Gradient Boosting Machine [42], and, as such, combines several regressors (frequently called *weak learners*) by averaging. In our case, regression trees are used as $N$ *weak learners* returning scalar sector count $h_i(\bar{x}_p)$ as a function of $p^{th}$ features' sample set $\bar{x}_p^T = [\bar{x}_t \quad \bar{x}_w]_p$, where

- $\bar{x}_t$ are time features: date and time (UTC).
- $\bar{x}_w$ are atmospheric weather data features: temperature, wind speed, wind direction, humidity, and pressure.
- $i$ is the index of an individual *weak learner* ($i = 1, \ldots, N$).
- $p$ is the index of an individual sample ($p = 1, \ldots, P$) belonging to the $P$-sized training set.

Each *weak learner* is chosen as a function which minimizes a least squares loss function. The algorithm may be understood as a two-staged (or *matching pursuit* [43]) optimization problem. In the first stage, an optimal *weak learner* for each regressor type is obtained. In the second stage, an optimal additive model is obtained by weighting the optimal *weak*

*learners* and producing therefore a scalar sector count estimator/predictor $F(\bar{x})$. This is done in a greedy fashion, [44]:
1. The estimator is initialized with the mean of the sector count training set ($\bar{y} = [y_1 \cdots y_P]^T$) as a best guess: $F_0 = \text{mean}(\bar{y})$.
2. A pseudo-residual sample ($\tilde{y}_p$) can now be computed like $\tilde{y}_p = y_p - F_{i-1}(\bar{x}_p)$ for the $i^{th}$ term of the additive model being trained.
3. The *weak learner* composing the next term is $h_i = \arg\min_h [\tilde{y}_1 \ldots \tilde{y}_P][\tilde{y}_1 \ldots \tilde{y}_P]^T$.
4. For $i=0,\ldots,N$: the ensemble receives the update according to the step length $\rho_i$: $F_i(\bar{x}) = F_{i-1}(\bar{x}) + \rho_i h_i(\bar{x})$.

This procedure is applicable for each dataset of a specific sector and the resulting predictors are now ready to be stored on the ML collection of the P-SWIM database. Fig. 4 illustrates the convergence of the Gradient Boosting Machine as new regression tree *weak learners* are added to the model. In this example just a few steps are used, in our application case $i = 1,\ldots,400$.

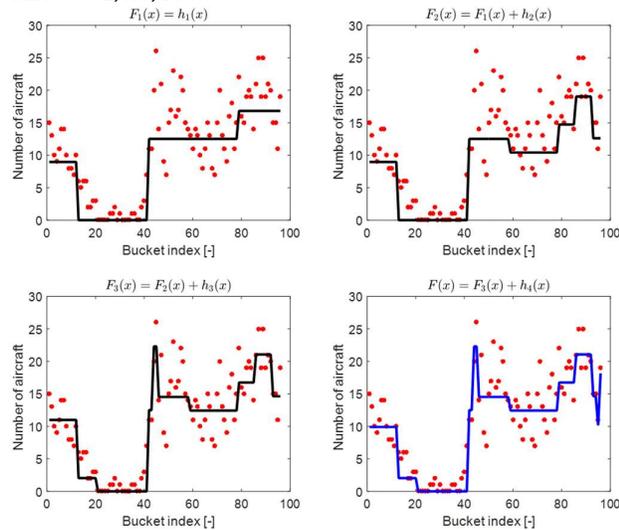

Fig. 4: Example of additive terms of the Gradient Boosting Predictor Model based on Decision Tree Weak Learners for a single day sector count training set. The dots represent the actual values and the solid line the predicted values.

### D. Model validation

It is desirable to have a training dataset with the largest size manageable by the available computing power, but it is equally desirable to have an idea on how accurate a prediction can be. For this purpose, we use the *k*-fold cross-validation approach, which subdivides the dataset into *k* subgroups of randomly chosen samples. One of these subgroups is used as validation set as the remaining are used as training set. This process is repeated *k* times while exchanging the subgroups which are used as training and validation data. At the end of the process the mean of the computed error metric is averaged to produce a performance index for the estimator numerically model based on that dataset.

The error measure used in our cross-validation effort was conceived to fairly gauge the accuracy according to the problem of sector count prediction. A data exploration on the processed data was carefully driven in order to determine a meaningful way of evaluating both highly populated sectors and those with sparse occupation. In Equation 1 below, $S_{sc}$ expresses the score used in the cross-validation of the sector count predictor using $N$ samples in the validation dataset. Individual samples $y_k$ belong to the sector count validation dataset array $y$; estimated values $F(\bar{x}_k)$ are a function of explanatory variables ($\bar{x}_k$) in the corresponding prediction domain.

$$S_{sc} = \frac{1}{N}\sum_{k=1}^{N} e^{-\frac{|y_k - F(\bar{x}_k)|}{\text{mean}(y)}} \quad \text{(Eq. 1)}$$

This metric converges to zero if predicted values are too far from the actual values. Otherwise, it converges to one in case of perfect prediction.

### E. Outlier rejection method

It was common to find entire sector count curves, i.e. a whole day, not describing usual aircraft count behavior. We did not investigate the causes of these anomalies, but we sought to have an automatic way to decide if a curve shall be accepted as a member of a training set or not. Our approach is to estimate a linear trend on a specific time bucket, say between 12:00 and 12:15, for all stored sector counts for a certain day in the week. The sample values selected to compose the trend look like those illustrated in Fig. 5.

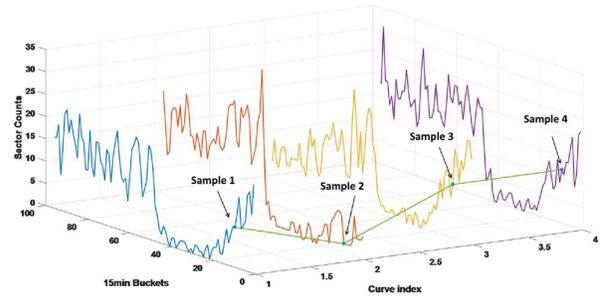

Fig. 5: Samples selected at the same bucket to compose a reference trend.

The trend is then compared against the sample values, with absolute error values summed up to compose scores for the individual curves. Fig. 6 shows the scores computed for all curves compared against the mean value of the scores, which is the rejection threshold. In this case, curves 3, 7, 10, and 15 are rejected because they exceed the threshold. In Fig. 7, the left-hand side plot shows the results with all curves, including the rejected ones, and the right-hand side plot shows the results with only those curves which are meaningful to be used in the training scheme.

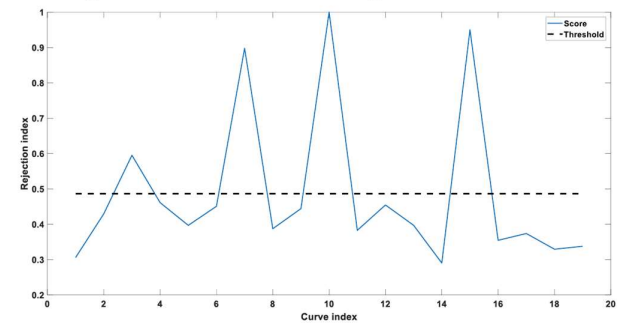

Fig. 6: Accumulated (and normalized) curve scores compared against mean (rejection threshold).

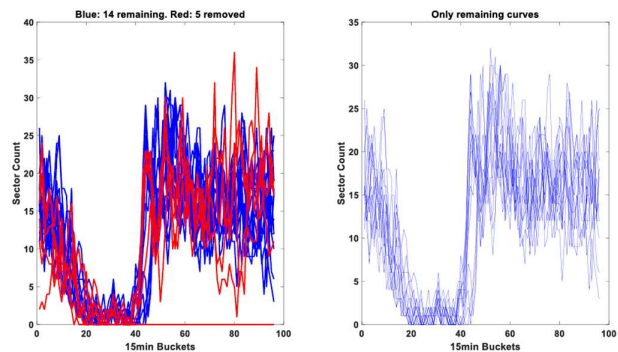

Fig. 7: Results from rejection scheme: all curves including rejected (left) and only those curves which are selected as meaningful for training (right).

### F. Criteria for evaluating model performance

In order to evaluate the performance of our aircraft count prediction scheme, three characteristics are of major importance: duration of training, response to user's prediction requests, and accuracy of predictions.

The **duration of training** is desired to be small, however a value in the order of a few hours is acceptable, since it can be done in less time than a data collection cycle, which is one day. At the end of a specific daily dataset a new processing routine starts to map new R-SWIM entries into P-SWIM. It is highly desirable to have predictors available in the current day with information gathered during the previous day. Thus, a good measure of effectiveness for the duration of training is the amount of time in the current day in which the predictors work without samples from the previous day. Optimal and sub-optimal values lie between 0h and 1h, while worst values between 23h and 24h. Currently, the training takes 0.91h (55min).

The **response time to user's prediction requests** is another time-based dimension of performance evaluation. Based on subjective perception of users, a limit value of 15s is specified. The microservices execute JSON message decoding of prediction request, preparation of query statement, query execution, and encoding of retrieved data into JSON format. In our tests, all these steps never surpassed 5s.

The **accuracy of predictions** is computed upon the predicted values under a cross-validation effort. The idea is to maximize the value expressed in Equation 1. This accuracy measure becomes the definition of the *prediction score* indicated in the examples below.

Both duration of training and response to user's prediction requests are criteria easily met by the framework. They rely heavily on the processing scheme explained previously. Without a suitable processing pipeline, it would not be possible to achieve the time values currently met.

The accuracy of predictions is of major concern and depends on the machine learning technique employed, in our case the gradient boosting. The size of the dataset contributes to the quality of predictions as well. In our case, the dataset is composed by 153 days. This means that the total number of messages accumulated in these days is organized in five groups to perform training and validation, i.e. cross-validation. On the other hand, the predictors used by the microservices always use the entire dataset for training while cross-validation is used only for purposes of performance evaluation. The next section summarizes results for the accuracy of predictions scored according to Equation 1 for cross-validation using the whole dataset.

### G. Results

The airspace sectors comprised in this case study are shown in Fig. 8, according to a color scale representing the predicted aircraft count for a certain period of time. The warmer colors (e.g. red, orange) represent the sectors with higher aircraft counts, while the colder colors (e.g. greens) represent sectors with low aircraft counts. This view is simplified, because there are sectors that overlap on the horizontal projection, but with distinct volumes as defined by their altitudes. The total number of sectors is 1534.

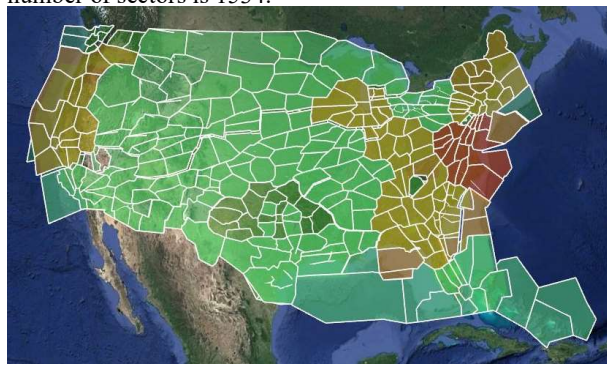

Fig. 8: Horizontal view of airspace sectors, with aircraft count represented by the color scale.

As some sectors barely experience a single flight in a day while others count to more than 2500 daily flights, figure Fig. 9 shows the scatter plot of accuracy scores achieved during cross-validation for each sector, with score values ordered as function of their respective sectors' daily counts. Moderately populated sectors (daily counts smaller than 500 flights) present an accuracy deterioration. One reason could be an uncertainty in the available data from R-SWIM database. It is known that there is lack of messages for some flights, which do not allow us to calculate sector count suitably. This means that the training dataset has an inherent noise which is difficult to characterize.

It is interesting to observe that sectors with low aircraft count have high accuracy, perhaps because the model achieves a high score by guessing that no flight at all will occur in each time window.

The approach adopted here was to investigate the messages used to compute the sector count and categorize them with a reliability measure. We noticed the following confusion cases: 1) A message with departure time is issued prior to the actual departure time, thus being only an estimated time; 2) A message indicating arrival time prior to a corresponding departure time; 3) A message with departure time issued after the message with arrival time; and 4) Other messages with different times for the same event. We though that these occurrences are somehow related to the uncertainty causing accuracy degradation. The rationale resides on the following desirable order of messages: `departureInformation`, `trackInformation`, `flightSectors`, and `arrivalInformation`. It is possible to compute sector count buckets with another message order, but this is the ideal one. Given an ambiguous pattern, different from the above, the possible hypotheses for the sequence of events are ordered from 1 to 3 and assigned to each time bucket, resulting in a feature called uncertainty measure. Using this feature to predict sector counts, one can achieve significantly higher scores, see the green scatter plot of Fig. 9, and corresponding average in blue.

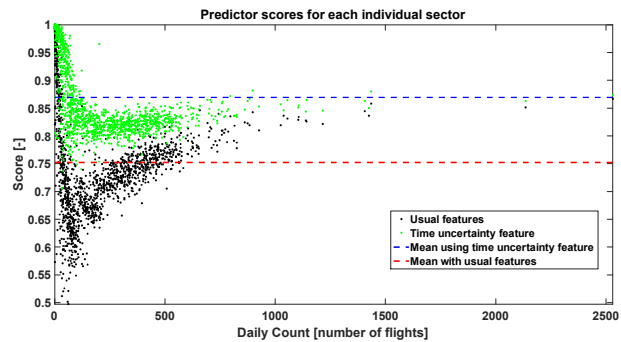

Fig. 9: Scores vs. sectors (ascending order of daily count) including additional feature.

There is a limitation in this uncertainty measure, since it could not be used in an on-line counting scheme before the flight finishes and all its messages are collected. However, it suffices for offline training of the model, and it points this research in the direction that other correlation features can be extracted from the same dataset. Increasing the size of the training dataset can also improve accuracy, but this can only be achieved in a fixed linear rate with more days of data collection (feeding R-SWIM).

*H. Use of Weather Variables*

Another analysis worth of reporting here is about the use of weather information. Weather has strong correlations with airport and airspace capacity, because it is easy to demonstrate the impact caused low visibility to airport capacity, and that storms cause to airspace sector capacity. However, the only weather information that could be univocally associated to a single sector, in the dataset analyzed, was that of MEteorological Terminal Air Report, or METAR. As the name implies, it is valid to a terminal, which means that it is collected nearby an airport. By directly correlation, only the airspace sectors that immediately contain an airport with a metrological station have METARs. Thus, just a minority of sectors had this information available. Notwithstanding, a comparison of prediction score of these sectors, with and without using METAR information, showed that there was no increase in prediction score with METAR. We did not deep dive to understand this unexpected result, however it may be that either there is not a sufficient number of METAR messages to cover all the time buckets, or that the model was not properly tuned for these additional variables.

## V. FINAL REMARKS

Our big data system is capable of ingesting large volumes of aeronautical data, collected from SWIM, and train Machine Learning models capable of predicting airspace sector congestion, by means of the aircraft count per sector, segmented in time buckets. Although the congestion of an airspace sector is more appropriately measured by taking into account the whole scene complexity, involving factors that affect the subjective difficulty perceived by a human Air Traffic Controller to perform its job [7], the aircraft count criterion is easier to attain and independent of expert opinion. However, the strength of our approach is to be virtually unlimited in the prediction horizon. Given a sector name, a time bucket in the future, defined by date and time, a fixed time step length, and a weather forecast for that time bucket, our models are capable of predicting the aircraft count for that sector with an average accuracy score of 87%, without looking at flight schedules. We believe that predictive capabilities like this can greatly improve the efficiency of ATM around the globe, by helping initiatives involving proactive management of capacity balance, such as dynamic resectorization [45], Collaborative Trajectory Option Set (CTOP) [46, 47], and several others.

Nevertheless, the accomplishments here reported are just a small demonstration of what can be achieved, and there are several directions of work that will help to improve our predictive capabilities, such as:

1. Upgrading the congestion measure to a workload difficulty measure based on a ponderation between total aircraft count and the traffic density per airspace volume.

2. Upgrading the congestion measure to an enhanced workload measure such as DD [7].

3. Use of published flight schedules.

4. Improving the use of weather information by including more comprehensive forecast data such as GFS [48], which is another big data source on its own, and doing more experiments to determine which weather variables are really indicative of changes in airspace capacity.

## VI. REFERENCES

[1] FAA, "Air Traffic by the Numbers," Aug 2022. [Online]. Available: https://www.faa.gov/sites/faa.gov/files/air_traffic/by_the_numbers/Air_Traffic_by_the_Numbers_2022.pdf. [Accessed 21 Jan 2023].

[2] International Air Transport Association (IATA), "Global Outlook for Air Transport," December 2022. [Online]. Available: https://www.iata.org/en/iata-repository/publications/economic-reports/global-outlook-for-air-transport---december-2022/. [Accessed Jan 2023].

[3] Boeing, "Commercial Market Outlook 2022-2041," [Online]. Available: https://www.boeing.com/commercial/market/commercial-market-outlook/index.page. [Accessed Jan 2023].

[4] ATAG, "Commitment to fly net zero 2050," 5 October 2021. [Online]. Available: https://aviationbenefits.org/media/167501/atag-net-zero-2050-declaration.pdf. [Accessed Jan 2023].

[5] ICAO, *Global Air Navigation Plan - Doc 9750,* 2016.

[6] J. A. Guttman, P. Kopardekar, R. H. Mogford and S. L. Morrow, "The Complexity Construct in Air Traffic Control: A Review and Synthesis of the Literature," 1995.

[7] I. V. Laudeman, S. Shelden, R. Branstrom and C. Brasil, *Dynamic Density: An Air Traffic Management Metric,* 1998.

[8] M. Albasman and J. Hu, "An Approach to Air Traffic Density Estimation and Its Application in Aircraft Trajectory Planning," in *Proceedings of the 24th Chinese Control and Decision Conference (CCDC)*, Taiyuan, China, 2012.

[9] P. Kopardekar and S. Magyarits, "Measurement and Prediction of Dynamic Density," in *5th USA/Europe ATM R&D Seminar*, Budapest, 2003.

[10] P. Kopardekar, A. Schwartz, S. Magyarits and J. Rhodes, "Airspace Complexity Measurement: An Air Traffic Control Simulation Analysis," *International Journal of Industrial Engineering,* vol. 16, no. 1, pp. 61-70, 2009.

[11] B. Sridhar, K. S. Sheth and S. Grabbe, "Airspace Complexity and its Application in Air Traffic Management," in *2nd USA/EUROPE AIR TRAFFIC MANAGEMENT R&D SEMINAR*, Orlando, FL, USA, 1998.

[12] A. J. Masalonis, M. B. Callaham and C. R. Wanke, "Dynamic Density and Complexity Metrics for Realtime Traffic Flow Management," in *5th USA/Europe ATM R&D Seminar*, Budapest, 2003.

[13] A. Klein, M. D. Rodgers and K. Leiden, "Simplified Dynamic Density: A Metric for Dynamic Airspace Configuration and NextGen Analysis," in *Proceedings of the IEEE/AIAA 28th Digital Avionics Systems Conference*, Orlando, FL, USA, 2009.

[14] C. F. Lai and S. Zelinski, "Simplified Dynamic Density Based Capacity Estimation," in *2009 IEEE/AIAA 28th Digital Avionics Systems Conference*, Orlando, FL.

[15] S. Ayhan, J. Pesce, P. Comitz, D. Sweet, S. Bliesner and G. Gerberick, "Predictive analytics with aviation big data," in *2013 Integrated Communications, Navigation and Surveillance Conference (ICNS)*, Herndon, VA, USA.

[16] Eurocontrol, "System-wide information management (SWIM)," [Online]. Available: https://www.eurocontrol.int/concept/system-wide-information-management. [Accessed 22 Dec. 2022].

[17] S. Ayhan and P. Comitz, "Swim interoperability with flight object mediation service," in *2009 IEEE/AIAA 28th Digital Avionics Systems Conference*, Orlando, FL, USA, 2009.

[18] S. Ayhan, P. Comitz and V. Stemkovski, "Aviation Mashups," in *2009 IEEE/AIAA 28th Digital Avionics Systems Conference*, Orlando, FL, USA, 2009.

[19] S. Ayhan and H. Samet, "Data Management and Analytcs System for Online Conformance Monitoring and Anomaly Detection," in *Proceedings of the 27th ACM SIGSPATIAL International Conference on Advances in Geographic Information Systems*, Chicago, IL, USA, 2019.

[20] FAA, "SWIM Services," [Online]. Available: https://www.faa.gov/air_traffic/technology/swim/products/. [Accessed Jan 2023].

[21] FAA, "TFMData Service," [Online]. Available: https://cdm.fly.faa.gov/?page_id=2288. [Accessed Jan 2023].

[22] C. Düntgen, T. Behr and R. Güting, "BerlinMOD: A Benchmark for Moving Object Databases," *The VLDB Journal,* vol. 18, no. 6, pp. 1335-1368, 2009.

[23] V. Botea, D. Mallett, M. A. Nascimento and J. Sander, "PIST: An Efficient and Practical Indexing Technique for Historical Spatio-Temporal Point Data," *Geoinformatica,* vol. 12, no. 2, pp. 143-168, 2008.

[24] P. Cudre-Mauroux, E. Wu and S. Madden, "TrajStore: An Adaptive Storage System for Very Large Trajectory Data Sets," in *2010 IEEE 26th Int'l Conference on Data Engineering (ICDE 2010)*, Long Beach, CA, 2010.

[25] D. Xie, F. Li, B. Yao, G. Li, L. Zhou and M. Guo, "Simba: Efficient In-Memory Spatial Analytics," in *Proceedings of the 2016 Int'l Conference on Management of Data*, San Francisco, CA, 2016.

[26] A. Eldawy and M. F. Mokbel, "SpatialHadoop: A MapReduce Framework for Spatial Data," in *Proceedings of the 2015 IEEE 31st Int'l Conference on Data Engineering*, Seoul, South Korea, 2015.

[27] B. Zheng, H. Wang, K. Zheng, H. Su, K. Liu and S. Shang, "SharkDB: An In-memory Column-oriented Storage for Trajectory Analysis," *World Wide Web,* vol. 21, no. 2, pp. 455-485, 2018.

[28] H. Tan, W. Luo and L. M. Ni, "CloST: A Hadoop-based Storage System for Big Spatio-temporal Data Analytics," in *Proceedings of the 21st ACM Int'l Conference on Information and Knowledge Management*, Maui, HI, 2012.

[29] X. Xie, B. Mei, J. Chen, X. Du and C. S. Jensen, "Elite: An Elastic Infrastructure for Big Spatiotemporal Trajectories," *The VLDB Journal,* vol. 25, no. 4, pp. 473-493, 2016.

[30] Q. Ma, B. Yang, W. Qian and A. Zhou, "Query Processing of Massive Trajectory Data Based on Mapreduce," in *Proceedings of the First Int'l Workshop on Cloud Data Management*, Hong Kong, China, 2009.

[31] J. Bao, R. Li, X. Yi and Y. Zheng, "Managing Massive Trajectories on the Cloud," in *Proceedings of the 24th ACM SIGSPATIAL Int'l Conference on Advances in GIS*, Buringame, CA, US, 2016.

[32] X. Ding, L. Chen, Y. Gao, C. S. Jensen and H. Bao, "UlTraMan: A Unified Platform for Big Trajectory Data Management and Analytics," *Proceedings of the VLDB Endowment,* vol. 11, no. 7, pp. 787-799, 2018.

[33] "Apache Spark," 2019. [Online]. Available: https://spark.apache.org/. [Accessed Jan 2023].

[34] "Apache GEODE," [Online]. Available: https://geode.apache.org/. [Accessed Jan 2023].

[35] "Apache Ignite," [Online]. Available: https://ignite.apache.org/. [Accessed Jan 2023].

[36] "IndexedRDD for Apache Spark," [Online]. Available: https://github.com/amplab/spark-indexedrdd. [Accessed Jan 2023].

[37] J. Ramnarayan, S. Menon, S. Wale and H. Bhanawat, "SnappyData: A Hybrid System for Transactions, Analytics, and Streaming: Demo," in *Proceedings of the 10th ACM Int'l Conference on Distributed and Event-based Systems*, Irvine, CA, 2016.

[38] S. Ayhan, G. Balvedi, P. Costas, I. A. Wilson, A. C. Leite and Í. Romani de Oliveira, "Systems, methods, and apparatus to improve aircraft traffic control". US Patent 2021/0312819 A1, 7 Oct 2021.

[39] "MongoDB home page," [Online]. Available: https://en.wikipedia.org/wiki/MapReduce. [Accessed Jan 2023].

[40] "MapReduce," [Online]. Available: https://en.wikipedia.org/wiki/MapReduce. [Accessed Jan 2023].

[41] P. Kopardekar and S. Magyarits, "Measurement and Prediction of Dynamic Density," in *Proceedings of the 5th USA/Europe ATM R&D Seminar*, Budapest, 2003.

[42] J. H. Friedman, "Greed function approximation: a gradient boosting machine," *Annals of statistics,* vol. 29, no. 5, pp. 1189-1232, 2001.

[43] S. G. Mallat and Z. Zhang, "Matching pursuits with time-frequency dictionaries," *IEEE Transactions on Signal Processing,* vol. 41, no. 12, pp. 3397-3415, 1993.

[44] M. Kuhn and K. Johnson, Applied Predictive Modeling, Springer, 2013.

[45] Í. Romani de Oliveira, R. J. G. Teixeira and P. S. Cugnasca, "Balancing the Air Traffic Control Workload Through Airspace Complexity Function," in *1st IFAC Symposium on Multivehicle Systems (MVS)*, Salvador, Brazil, 2006.

[46] "Collaborative Trajectory Options Program (CTOP)," [Online]. Available: http://www.nbaa.org/ops/airspace/tfm/tools/ctop.php. [Accessed 15 Jan 2023].

[47] L. Cruciol, J.-P. Clarke and L. Weigang, "Trajectory Option Set Planning Optimization under Uncertainty in CTOP," in *Proceedings of the 2015 IEEE 18th International Conference on Intelligent Transportation Systems*, Gran Canaria, 2015.

[48] NOAA, "Global Forecast System," [Online]. Available: https://www.ncei.noaa.gov/products/weather-climate-models/global-forecast. [Accessed 18 Dec 2022].